# Integrating Symbolic Natural Language Understanding and Language Models for Word Sense Disambiguation

**Kexin Zhao**  KEXINZHAO2029.1@U.NORTHWESTERN.EDU
**Ken Forbus**  FORBUS@NORTHWESTERN.EDU
Qualitative Reasoning Group, Northwestern University, 2233 Tech Drive, Evanston, IL 60208 USA

## Abstract

Word sense disambiguation is a fundamental challenge in natural language understanding. Current methods are primarily aimed at coarse-grained representations (e.g. WordNet synsets or FrameNet frames) and require hand-annotated training data to construct. This makes it difficult to automatically disambiguate richer representations (e.g. built on OpenCyc) that are needed for sophisticated inference. We propose a method that uses statistical language models as oracles for disambiguation that does not require any hand-annotation of training data. Instead, the multiple candidate meanings generated by a symbolic NLU system are converted into distinguishable natural language alternatives, which are used to query an LLM to select appropriate interpretations given the linguistic context. The selected meanings are propagated back to the symbolic NLU system. We evaluate our method against human-annotated gold answers to demonstrate its effectiveness.

## 1. Introduction

Word sense disambiguation is one of the fundamental challenges in natural language understanding. When humans encounter an ambiguous word in a sentence, we can easily use context to determine the intended meaning. However, AI systems often struggle with such tasks, especially when dealing with fine-grained semantic distinctions that go beyond basic category identification.

Many current disambiguation approaches focus on the coarse-grained level, such as identifying broad semantic frames or synset categories (Navigli, 2009). While they perform well in basic language processing tasks, they are insufficient when deeper semantic understanding is required. Fine-grained disambiguation, which identifies subtle differences within the same semantic category, is still mostly unsolved because annotated training data is limited and making precise semantic distinctions is inherently difficult (Navigli, 2009). For example, for the sentence "The traffic light turned yellow," the verb "turn" can be easily identified as describing a change, but fine-grained disambiguation must determine whether this represents an external or an internal change.

Current methods face a core trade-off. Symbolic approaches offer structured and interpretable outputs but have difficulty handling linguistic variation and flexible contexts. Neural approaches manage context effectively but produce opaque decisions and lack the structured representations needed for advanced reasoning. Both approaches typically depend on large amounts of hand-annotated training data, making fine-grained distinctions especially costly to handle. For example, the NextKB ontology[1] we use has over 80,000 concepts and 20,000 relations. Creating a training set for it would be a massive undertaking.

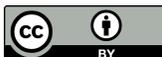

This paper presents a hybrid approach that combines the complementary strengths of symbolic natural language understanding systems and large language models. Our work was inspired by the idea that if a formal reasoning system can generate natural language representations of its reasoning and outputs, then this creates a powerful capability that the system can communicate with Large Language Models using natural language as their API (Shepard, 2025). Our method uses symbolic systems to generate structured candidate meanings and take advantage of the contextual understanding ability of large language models to select appropriate interpretations, without relying on hand-annotated training data. The approach achieves consistent disambiguation accuracy at both coarse and fine-grained semantic levels, demonstrating the effectiveness of integrating symbolic and neural methods for word sense disambiguation.

## 2. Background

### 2.1 CNLU

Companions Natural Language Understanding (CNLU; Tomai & Forbus, 2009) is a rule-based semantic parser designed to process natural language as part of the Companions cognitive architecture (Forbus & Hinrichs, 2017). As shown in Figure 1, the system builds upon a version of Allen's (1994) TRAINS parser, modified so that it can query the knowledge base dynamically and call on other reasoning services during the parsing process. To handle semantic interpretation,

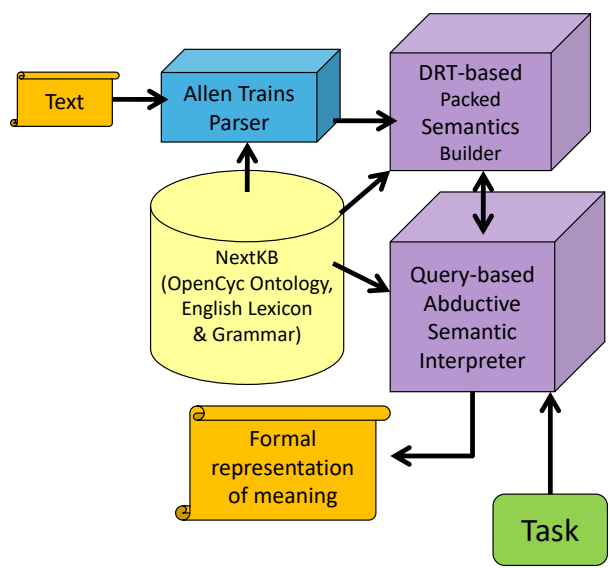

*Figure 1*. CNLU overall architecture.

---

[1] https://www.qrg.northwestern.edu/nextkb/index.html.

CNLU maps linguistic input to the OpenCyc ontology through FrameNet as a bridge. The system uses Discourse Representation Theory (Kamp & Reyle, 1993) to process sophisticated linguistic phenomena including complex conditionals, quantification, quotation, and counterfactual statements. It converts everyday text into structured representations that computers can use for reasoning. The system uses a symbolic approach to interpret language, applying grammar rules and a large scale lexicon from the knowledge base to analyze its meaning. CNLU breaks down input text into logic-based semantic structures with precise and explicit meaning, making it easier for automated systems to reason with the output. One major advantage is that humans familiar with CNLU's ontology can easily read and understand what the system produces. Figure 2 shows an example of an CNLU generated the structured representation for the phrase "my parents." Here, symbols with numerical suffixes such as "parent4172" are automatically generated unique discourse identifiers that the system assigns to distinguish between different entities.

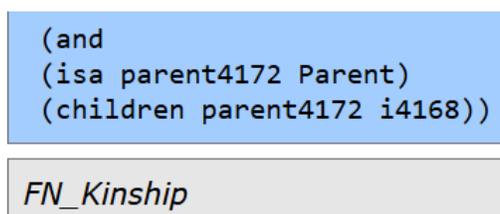

*Figure 2*. CNLU generated representation for the phrase "my parents."

CNLU creates semantic representations that are much more detailed and structured than standard resources like WordNet or FrameNet[2]. It generates semantic interpretations that are grounded in the NextKB ontology, which is a superset of the OpenCyc ontology. The system identifies complex and subtle relationships between words, categorizes various types of concepts, and recognizes the corresponding semantic constraints. This provides a solid foundation for complex language understanding and reasoning tasks, though it also makes the processing more challenging. When CNLU encounters words or phrases with multiple possible meanings, it generates candidate interpretations for each ambiguity. These candidates are organized into choice sets, with each option representing a different way to understand the meaning. The system keeps all possibilities open until it can gather enough context to make final decisions. Therefore, since CNLU works with such a rich and complex semantic space, accurate word sense disambiguation remains a significant challenge.

## 2.2 Coarse-Grained vs. Fine-Grained

Word sense disambiguation can work at different levels of detail (Navigli, 2009). Coarse-grained disambiguation focuses on making broad distinctions between word meanings. It typically works at the level of semantic frames, like those found in WordNet or FrameNet. For example, the word

---

[2] https://framenet.icsi.berkeley.edu/

K. Zhao and K. Forbus

"run" can be disambiguated into several clear categories: self-motion (e.g. the children run home), operation (e.g. a computer program runs), or leadership (e.g. someone runs for office). These distinctions are relatively easy to understand and separate from each other. Most existing disambiguation systems work at this coarse level because the categories are well-defined and training data is available. On the other hand, fine-grained disambiguation goes much deeper. It makes subtle distinctions within the same semantic frame or category. While coarse-grained systems might identify that "run" refers to self-motion, fine-grained systems would further specify what type of motion is involved, who runs, and what is the purpose. Knowledge bases like OpenCyc and NextKB provide these detailed semantic distinctions that go far beyond what FrameNet offers. For example, as shown in Figure 4, for the word "turn," both FrameNet and NextKB will provide one semantic frame as Undergo_change, but NextKB can further distinguish between different types of change within the same frame.

The difference between coarse and fine-grained disambiguation approaches affects what kind of reasoning systems can do. Coarse-grained representations work well for basic language understanding tasks. They can handle common disambiguation problems and provide reasonable semantic interpretations. Fine-grained representations enable more sophisticated reasoning because they capture subtle differences in meaning that matter for complex inference tasks. For example, skiing versus snowboarding both fit within FrameNet's self-motion frame, but they require different equipment and skills. Fine-grained disambiguation faces unique challenges because, unlike coarse-grained approaches, there are fewer available resources and training datasets for it. The semantic distinctions are often too nuanced and require deeper understanding of context and world knowledge. However, systems that can perform fine-grained disambiguation have access to much richer semantic information for reasoning and learning.

## 3. Method

We present a hybrid approach that combines symbolic natural language understanding systems with language models for word sense disambiguation.

### 3.1 Overall Framework

Our approach works through four steps, as shown in Figure 3. First, a symbolic NLU system analyzes the input sentence and generates multiple candidate meanings for ambiguous words. Second, for each ambiguous word, its symbolic representations are converted into natural language descriptions that language models can understand. Third, we query a language model to select the most appropriate meaning given the context. Finally, we integrate the selected meaning back into the symbolic NLU system and apply the corresponding semantic information.

### 3.2 Symbolic Candidate Generation

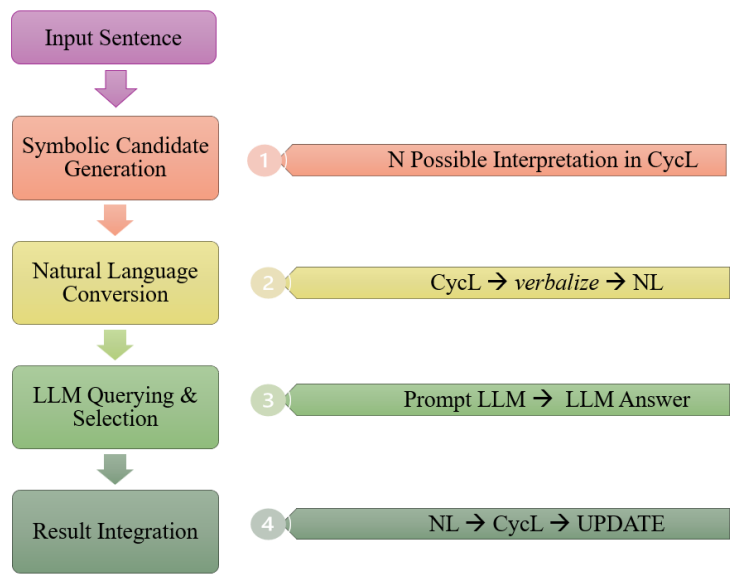

*Figure 3*. Overall framework of the hybrid word sense disambiguation approach.

The symbolic NLU system first performs syntactic and semantic analysis on the input sentence. For each ambiguous word, the system retrieves multiple possible meanings from its knowledge base. These candidate meanings are represented as complex logical forms with detailed semantic and relational information.

Consider the sentence "The traffic light turned yellow" with the ambiguous verb "turn." The system generates eighteen different candidate meanings. Figure 4 shows three representative examples that illustrate fine-grained distinctions within the same frame as well as coarse-grained difference across frames:

```
Candidate 1:
     (and (isa turn38450 TurningSomethingIntoSomethingElse)
          (objectActedOn turn38450 traffic-light38442))
     Frame: FN_Undergo_change
Candidate 2:
     (and (isa turn38450 IntrinsicStateChangeEvent)
          (objectActedOn turn38450 traffic-light38442))
     Frame: FN_Undergo_change
Candidate 3:
     (and (isa turn38450 SubmittingSomething)
          (relationInstanceExists infoTransferred turn38450 Document))
     Frame: FN_Submitting
```

*Figure 4*. Three candidate meanings for the ambiguous verb "turn."

Each candidate provides a structured semantic representation that captures different aspects of the possible meaning. This includes ontological classifications that specify the type of action or entity, relational predicates that define how elements connect to each other, and constraints that determine how the word relates to other elements in the sentence. The system can generate many candidates for a single ambiguous word, each representing a different possible interpretation. In the examples below, the first candidate treats "turn" as an external transformation that turns one thing into another, the second as an internal state change, while the third belongs to a completely different frame and represents document submission.

When a sentence contains multiple ambiguous words, the system processes them one at a time instead of simultaneously. This makes sure that once a word is disambiguated, the remaining ambiguous words are constrained to semantically compatible choices, preventing conflicting interpretations within the same sentence.

### 3.3 Natural Language Conversion

To make complex symbolic candidate meanings understandable to language models, we use FIRE's built-in verbalize function. Verbalize converts CycL expressions into English text (Nakos, Demel, & Forbus, 2025; Wilson et al., 2019). This approach of converting formal logical representations into natural language is similar to template-based methods developed for systems like Cyc where assertions are turned into readable English (Baxter et al., 2005). The function uses predefined text templates and simple linguistic information to generate output.

For each candidate meaning, we extract the individual logical statements and call the verbalize function on each one. The conversion process has three steps. We extract expression components by separating individual sub-expressions from the conjunctive structure. We then convert each sub-expression individually by calling verbalize on each part. Finally, we combine the resulting natural language fragments into a complete meaning description using semicolons as separators. Using our previous examples, the conversion works as shown in Figure 5.

> **Candidate 1 converts to:** "turn is a turning; traffic light is acted on during turned."
>
> **Candidate 2 converts to:** "turn is a becoming; traffic light is acted on during turned."
>
> **Candidate 3 converts to:** "turn is a submitting."

*Figure 5*. Three candidate meanings conversion to natural language.

This conversion mechanism transforms complex symbolic representations into natural language options that language models can understand and process. The verbalized descriptions capture the essential semantic differences between candidates while remaining understandable to both the language model and humans, though the verbalization process may not always produce the perfect outputs.

### 3.4 LLM Querying and Selection

We build standardized prompt templates for the language model. The prompt includes the original sentence, identifies the ambiguous word, and presents all candidate meanings as numbered options. Figure 6 shows an example of the specific prompt sent to the large language model.

> Sentence: "The traffic light turned yellow." Please select the most appropriate meaning for the word "**turn**"
> Options:
> 1. turn is a turning; traffic light is acted on during turned
> 2. turn is a becoming; traffic light is acted on during turned
> 3. turn is a submitting
> ...
> [the rest of the options]
> Reply only with the option number.

*Figure 6*. Example prompt template for LLM.

We query the language model for the most likely interpretation of the word using the current sentence as context. The model returns an option number which gets parsed and mapped back to the corresponding symbolic meaning representation. We note that for sentences in connected text including a small set of prior sentences as context will almost certainly be beneficial, but for sentences drawn from a benchmark, such sentences are often not available, and hence we only used the current sentence as context for uniformity.

### 3.5 Result Integration

The selected meaning gets integrated back into the symbolic NLU system through several steps. We map the chosen option number back to the full symbolic representation, including all logical predicates. The selected results are then propagated to the rest of that sentence's analysis via a truth maintenance system. Once a choice for one word is made, the system immediately updates the discourse state and rules out parse trees that are incompatible with the confirmed selection, thus maintaining consistency across the entire sentence's semantic analysis. Additionally, this propagation allows the system to automatically access the corresponding coarse-grained semantic frame information through pre-established bindings in the knowledge base based on the selected fine-grained candidate.

The method successfully bridges symbolic and neural approaches by converting complex logical representations into natural language that LLMs can process. This enables the system to make fine-grained semantic distinctions without requiring hand-annotated training data. Moreover, the selected meanings maintain their original symbolic structure, allowing seamless integration back into the NLU system for further processing.

### 4. Experiments and Results

### 4.1 Dataset and Gold Standard

We evaluate our approach on premises from the Choice of Plausible Alternatives (COPA) dataset (Roemmele, Bejan, & Gordon, 2011). COPA premises contain rich semantic content with frequent lexical ambiguities that make them well-suited for testing both coarse and fine-grained disambiguation systems. These sentences often describe causal relationships and human actions, leading to ambiguous interpretations of key predicates. For example, in "My body cast a shadow over the grass," the verb "cast" could refer to physical projection or theatrical performance, while "The woman tolerated her friend's difficult behavior" contains "tolerated" which might indicate emotional endurance or physical resistance.

Our test set consists of 50 COPA premise sentences[3] containing 114 ambiguous words that require disambiguation. We chose these sentences to span various semantic domains and include both verb and noun ambiguities. Each sentence averages about 2.3 ambiguous terms, providing sufficient complexity to test our method's effectiveness across different linguistic contexts.

We constructed the gold standard through human annotation by the first author, where each ambiguous word was manually assigned the correct predicate and semantic frame based on the sentence context. The annotation process followed consistent criteria by considering both the immediate syntactic context and the broader discourse meaning of each sentence. We acknowledge that our single-annotator evaluation conducted by the primary researcher introduces potential bias into the experiment, and it would be preferable to use the entire dataset. However, these initial results already provide evidence for the approach, which we plan to confirm with independent annotations and larger datasets.

### 4.2 Experiment Setup

We use the Microsoft Phi4 LLM (Abdin et al., 2024) here and use BERT (Devlin et al., 2018) as a baseline. We evaluate performance at two levels of semantic granularity. The BERT Frame Classifier, our primary baseline, uses a pre-trained BERT model fine-tuned for FrameNet frame classification (Nakos & Forbus, 2023). Given a sentence and target word position, this classifier outputs probability distributions over candidate frames and selects the highest-scoring option. While useful for frame-level disambiguation, the classifier is coarse-grained and cannot distinguish between fine-grained choices within the same frame. It also cannot disambiguate choices without a FrameNet frame, and the conceptual coverage of the OpenCyc ontology is much broader than FrameNet.

To assess our approach's performance on fine-grained disambiguation capabilities, we also implement a BERT Random baseline that combines the BERT frame classifier with random predicate selection. This method first applies BERT to identify the correct FrameNet frame, when one exists. If the frame prediction is incorrect, the instance is marked as wrong. When the frame is correctly identified, the system randomly selects one predicate from within that frame. This baseline represents the limitation of systems that can only perform coarse-grained disambiguation.

---

[3] The full COPA consists of 1,000 sentences. We selected a tractable subset to ensure high-quality gold annotations, with large-scale studies planned for future work.

Our experiment design includes two evaluation settings. The coarse-grained evaluation measures accuracy at the frame level, comparing BERT's frame classification against our approach on the Phi4 for frame identification. The fine-grained evaluation assesses predicate-level accuracy, contrasting BERT Random with our complete Phi4-based disambiguation system.

We use accuracy as our primary evaluation metric, measuring the percentage of ambiguous words correctly disambiguated at each granularity level. The Phi4 model operates with default parameters, processing natural language descriptions of semantic alternatives to make disambiguation decisions based on sentence context.

### 4.3 Results

Our results demonstrate significant improvements for the Phi4-based approach across both evaluation settings. Table 1 shows the accuracy scores for each method at different levels of semantic granularity.

*Table 1*. Accuracy for our approach on Phi4 and the baseline BERT approach across two granularity levels.

|  | Coarse-grained | Fine-grained |
| --- | --- | --- |
| BERT | 69.3% | 20.2% |
| CNLU+Phi4 | **84.2%** | **82.5%** |

At the coarse-grained level, our approach on Phi4 achieves 84.2% accuracy compared to BERT's 69.3%. This suggests that our hybrid approach can effectively handle frame identification tasks.

The fine-grained results reveal even more dramatic differences. Our complete system achieves 82.5% accuracy for predicate-level disambiguation, while the BERT Random baseline manages only 20.2%. This gap underscores both the inherent difficulty of fine-grained disambiguation and the effectiveness of our approach in choosing precise predicate-level interpretations.

A particularly notable finding is the consistency of our method's performance across granularity levels. The difference between coarse-grained (84.2%) and fine-grained (82.5%) accuracy is very close, suggesting that our approach can simultaneously handle both frame identification and predicate selection with similar accuracy. This consistency contrasts with our BERT baseline, which shows a significant performance drop from coarse-grained (69.3%) to fine-grained (20.2%) disambiguation accuracy.

Error pattern analysis reveals that among the 96 cases where Phi4 correctly identified semantic frames, 94 achieved accurate predicate-level disambiguation. This finding provides important insights into our method's disambiguation capabilities as discussed in the next section.

The BERT Random baseline's poor performance highlights a fundamental limitation of approaches that rely solely on coarse-grained disambiguation. Even when frame identification succeeds, random selection among fine-grained alternatives will generate results barely above chance level, which does not inherently help with disambiguation at a deeper level.

## 5. Discussion

Our hybrid approach demonstrates the potential of combining symbolic natural language understanding systems with large language models for fine-grained word sense disambiguation. The experiment results reveal several important insights about how this integration performs across different levels of semantic granularity and what this means for practical natural language processing applications.

### 5.1 Strength

The elimination of training data requirements represents a significant contribution. Traditional disambiguation systems depend heavily on manually annotated datasets, which are expensive to create and are often domain specific. Our method avoids this limitation by using the symbolic system's candidate generation capabilities combined with the language model's semantic understanding. This method makes the system immediately applicable to new domains and languages, without the need for supervised training on annotated examples. The cost is the construction of a rule-based natural language generation capability (here, verbalize). In our experience, this cost is not high, because the goal is to find one good linguistic realization of a concept, versus understanding, where coverage means finding all ways that the concept might be expressed in language.

As presented above, the error pattern provides an important insight into how our approach handles semantic disambiguation. When our method successfully identifies semantic frames, it demonstrates strong reliability in fine-grained predicate selection. This concentrated effectiveness within correct conceptual boundaries suggests that our method engages meaningfully with deeper semantic distinctions and predicate-level reasoning.

Beyond its theoretical significance, this pattern offers crucial practical value for semantic processing applications. Frame-level reliability creates a stable foundation for fine-grained disambiguation, ensuring that predicate selection occurs within appropriate semantic boundaries. This consistency reduces the risk of major semantic misunderstandings while confining errors to manageable distinctions between related concepts. By operating within correct semantic frames, our approach enables more precise processing than existing coarse-grained methods, which often struggle to address the nuanced predicate distinctions that applications require.

### 5.2 Case Studies and Examples

#### 5.2.1 Fine-Grained Disambiguation

Fine-grained disambiguation reveals its value in cases where coarse-grained methods seem successful but miss important semantic distinctions. Returning to our previous example, consider "The traffic light turned yellow." For the word "turn," multiple frames are possible, including *FN_Undergo_change*, *FN_Submitting*, *FN_Change_direction*, etc. Both our approach and BERT correctly identify *FN_Undergo_change* as the appropriate frame. However, this frame contains multiple predicate interpretations with different meanings. One represents external transformation *(TurningSomethingIntoSomethingElse)*, while another represents internal state change *(IntrinsicStateChangeEvent)*.

This distinction matters for accurate understanding. A traffic light changing color is an intrinsic state change, not a transformation of one object into another. If there is further reasoning based on this, it can affect the understanding more severely, such as causality, object properties, and event types. For instance, if "turning yellow" was mistakenly identified as an external transformation, it would imply that the current traffic light no longer exists, and some agent caused such replacement. This may lead to inconsistencies when integrating new information about the current traffic light or attempting to find the nonexistent external agent, and ultimately to a failure to grasp the fundamental cycling behavior of a traffic light's color change.

*5.2.2 Coarse-Grained Disambiguation*

Our approach demonstrates strong performance in frame-level disambiguation. Consider the sentence "My body casts a shadow over the grass." For the word "grass," CNLU generates candidate meanings across different frames, which are then verbalized into natural language descriptions. The language model correctly identifies *(isa grass10581 Grass-Plant)* within the *FN_Plants* frame, because the surrounding physical context "casting a shadow" indicates the botanical meaning rather than alternative interpretations. In contrast, the BERT baseline incorrectly categorizes the same word under the *FN_Intoxicants* frame, indicating that it can make mistakes even at a broad semantic level.

*5.2.3 Information Completeness and Semantic Richness*

Beyond correct disambiguation, our method also demonstrates the ability to select semantically richer representations when multiple valid options exist within the same frame. Consider the sentence "The customer filed a complaint with the store manager." For the word "complaint," both approaches correctly identify the *FN_Complaining* frame, but multiple predicate-level representations are available. Our system selects the more informative option: *(and (isa complaint57061 Complaint) (recipientOfInfo complaint57061 store-manager57126))* rather than the simpler *(isa complaint57061 Complaint)*. This choice captures not only the existence of a complaint but also the key character, which is the recipient in this case, and their relationship explicitly mentioned in the sentence. This demonstrates our method's preference for semantically richer representations that preserve explicit relational information from the input.

*5.2.4 Failure Analysis*

While our approach achieves strong overall performance, we also want to examine its failures because they provide important insights into its limitations. These errors generally fall into two categories: minor predicate-level mistakes within correct frames, and incorrect frame identification.

The first category is illustrated by the sentence "The gardener wanted his plants to flourish." For word "flourish," our system correctly identifies the *FN_Thriving* frame but selects *(isa*

*flourish37584 GainingInWealth)* instead of the correct predicate focusing on plant growth events. This error demonstrates that even within appropriate semantic frames, the language model can sometimes be influenced by certain semantic connections rather than domain-specific interpretations suggested by the context. Such errors, while relatively infrequent, highlight the ongoing challenge of ensuring that language models fully integrate
contextual information when making fine-grained semantic distinctions.

More concerning are occasional failures where the system makes fundamental conceptual errors. In the sentence "The vandals threw a rock at the window," our approach incorrectly selects *(isa window40758 ComputerDisplayWindow)* instead of the correct *(isa window57271 WindowPortal)* within another frame. This error represents a serious misunderstanding of the physical context, confusing a building's window with a computer interface element despite clear indicators of physical vandalism in the sentence.

This type of error reveals a critical limitation: language models can make choices that violate basic physical world understanding. Such failures underscore the importance of developing robust validation mechanisms and highlight areas where purely neural approaches to semantic selection may require additional constraints or verification steps. Table 2 summarizes the distribution of error types across all misclassifications of the proposed system.

*Table 2*. Error analysis breakdown.

| Error Type | Percentage |
|---|---|
| Physical context | 30% |
| Action | 30% |
| Social & role | 15% |
| Emotion | 15% |
| State change | 10% |

Despite these limitations, our analysis shows that when frames are correctly identified, our approach demonstrates reliable fine-grained disambiguation. The method shows strong overall performance, especially compared with the baseline.

## 5.3 Limitations

While we present the above strong performance of our approach across two disambiguation tasks, there are three limitations.

Firstly, our evaluation relies on human-annotated gold standards, which may introduce subjectivity in the assessment. Fine-grained semantic distinctions can sometimes be interpreted differently by different annotators, especially when dealing with subtle predicate-level differences within the same semantic frame. To establish more valid and robust results, we will conduct future experiments with multiple annotators with inter-annotator agreement measurements, as well as testing on a larger dataset.

Secondly, our approach is implemented within our CNLU system. While the core methodology of using language models to select among symbolically generated candidates is generally applicable, the specific performance we observe may be influenced by the strengths and limitations of our particular symbolic system. The method's effectiveness in other symbolic frameworks or with different knowledge representation schemes would require additional validation and potentially some adaptation of the candidate generation and verbalization processes. However, the core requirements for adaptation are straightforward: any symbolic system that can generate distinguishable semantic alternatives for ambiguous expressions and can convert its internal representations into natural language descriptions can adopt our approach. For example, systems like Cyc that already possess similar capabilities could readily adopt it (Baxter et al., 2005). Therefore, this dependency does not invalidate our core contribution but suggests that implementation details may need adjustment for different symbolic NLU systems.

Thirdly, our method's performance depends on the quality of verbalizations. While our current approach works well as shown in the evaluation, different verbalization strategies could impact disambiguation accuracy. Future work should explore systematic approaches to improve this aspect.

Despite these limitations, our results demonstrate that the approach of combining symbolic candidate generation with neural selection mechanisms offers a promising direction for fine-grained word sense disambiguation that addresses key weaknesses in existing methods.

## 6. Related Work

### 6.1 LLM-only Disambiguation

Recent word sense disambiguation research increasingly uses large language models to resolve semantic ambiguity. These approaches typically use prompting strategies to have models select appropriate meanings from predefined options or generate natural language explanations for word senses in context (Sumanathilaka, Micallef, & Hough, 2024). Some methods fine-tune language models specifically for disambiguation tasks, while others rely on the models' inherent linguistic knowledge through carefully designed prompts (Yae et al., 2025).

Large language models process disambiguation tasks by drawing on patterns learned from extensive data from the internet during training (Brown et al., 2020). They can model contextual relationships within sentences and paragraphs, allowing them to consider broader linguistic context when making word sense selections. The models typically work with natural language descriptions of word meanings, which allows for flexible representation of semantic concepts without requiring rigid categorical schemes.

However, LLM-based approaches usually generate outputs in free natural language form, which requires additional processing when integrating with formal symbolic reasoning systems for further downstream tasks.

### 6.2 Traditional Symbolic Disambiguation

Traditional word sense disambiguation systems have relied on symbolic approaches that use explicit knowledge representations or hand-written rules to resolve semantic ambiguity. These methods typically use lexical resources such as WordNet, semantic networks, and domain-specific ontologies to provide structured representations of word meanings. Classic approaches include preference-based matching with semantic templates (Wilks, 1975) or selection preference constraints (Resnik, 1997) to identify appropriate word senses.

Specifically, some research has focused on such computational cognitive models for natural language understanding and disambiguation. Mcshane and Nirenburg (2021) have proposed a method centered on a primarily symbolic system which uses a rich semantic ontology. It uses multiple reasoning strategies to parse text and handles ambiguity by maintaining many candidate interpretations that are progressively refined through their designed six stages. Similar to our work, their approach builds upon a knowledge-rich symbolic system to generate structured semantic representations. However, our system is different from theirs. They follow a knowledge-based symbolic path where disambiguation relies on hand-crafted ontology, lexicon, and episodic memory, while our method combines symbolic NLU system with large language models.

The primary strength of symbolic approaches is their transparency. These systems provide clear explanations for their disambiguation choices and consistently produce the same outputs for identical inputs. The explicit semantic representations integrate naturally with other symbolic reasoning components, making them suitable for applications requiring structured knowledge manipulation.

Some analogical approaches have been explored that use human-like analogical reasoning and processing to perform disambiguation (Barbella & Forbus, 2013). One method is to construct cases from structured representations and analogically reason through and find the matching prior contexts for making word sense disambiguation.

However, symbolic systems face significant limitations in handling the flexibility and variability of natural language. They struggle with complex sentence structures, informal expressions, or linguistic patterns that fall outside their predefined rules. The manual construction of comprehensive rule sets requires extensive effort from domain experts, and these systems are difficult to adapt to new domains without substantial work. Additionally, their coverage depends heavily on the completeness of underlying knowledge resources, which limits their applicability to broad, real-world language processing tasks.

## 7. Conclusion

This work presents a novel hybrid approach that uses natural language descriptions as a direct interface between symbolic semantic representations and language model capabilities. This design addresses the complementary weaknesses found in pure approaches. Unlike purely symbolic systems, it handles linguistic variation through the language model component. Unlike pure neural approaches, it produces structured semantic outputs and maintains transparent decision processes. The natural language verbalization process enables this integration without modifying either component. The approach works particularly well for applications requiring semantic precision beyond what current coarse-grained methods can provide, while maintaining the linguistic flexibility that traditional symbolic systems lack.

There are two future directions to explore, in addition to larger-scale experiments. First, we plan to extend the hybrid methodology to co-reference resolution, which could address current limitations in linking pronouns and entities across sentences. Second, we aim to apply similar techniques to parsing disambiguation. Together, these extensions would demonstrate a broader applicability of symbolic–neural integration approach in natural language understanding tasks.

## Acknowledgements

This research is sponsored by the Office of Naval Research and Air Force Office of Scientific Research.